\documentclass[runningheads]{llncs}
\usepackage{subfigure}

\usepackage[T1]{fontenc}
\usepackage{amsmath,amssymb,amsfonts}
\usepackage{hyperref}
\usepackage{graphicx}
\begin{document}
\title{Swin Transformer for Robust Differentiation of Real and Synthetic Images: Intra- and Inter-Dataset Analysis}
%
%\titlerunning{Abbreviated paper title}
% If the paper title is too long for the running head, you can set
% an abbreviated paper title here
%
\author{Preeti Mehta\inst{1}\orcidID{0000-0002-6153-2145} \and
Aman Sagar\inst{2} \and
Suchi Kumari\inst{2}\orcidID{0000-0002-6748-5028}}
\authorrunning{M. Preeti et al.}
% First names are abbreviated in the running head.
% If there are more than two authors, 'et al.' is used.
%
\institute{National Institute of Technology Delhi, India \\
\email{preetimehta@nitdelhi.ac.in}\\
\and
Shiv Nadar Institute of Eminence, Delhi-NCR, India\\
\email{as624@snu.edu.in}, \\ \email{suchi.singh24@gmail.com} (Corresponding Author)}
\maketitle              % typeset the header of the contribution
\begin{abstract}
\textbf{Purpose}
This study aims to address the growing challenge of distinguishing computer-generated imagery (CGI) from authentic digital images in the RGB color space. Given the limitations of existing classification methods in handling the complexity and variability of CGI, this research proposes a Swin Transformer-based model for accurate differentiation between natural and synthetic images.

\textbf{Methods}
The proposed model leverages the Swin Transformer's hierarchical architecture to capture local and global features crucial for distinguishing CGI from natural images. The model's performance was evaluated through intra-dataset and inter-dataset testing across three distinct datasets: CiFAKE, JSSSTU, and Columbia. The datasets were tested individually (D1, D2, D3) and in combination (D1+D2+D3) to assess the model's robustness and domain generalization capabilities.

\textbf{Results}
The Swin Transformer-based model demonstrated high accuracy, consistently achieving a range of 97-99\% across all datasets and testing scenarios. These results confirm the model's effectiveness in detecting CGI, showcasing its robustness and reliability in both intra-dataset and inter-dataset evaluations.

\textbf{Conclusion}
The findings of this study highlight the Swin Transformer model's potential as an advanced tool for digital image forensics, particularly in distinguishing CGI from natural images. The model's strong performance across multiple datasets indicates its capability for domain generalization, making it a valuable asset in scenarios requiring precise and reliable image classification.

\end{abstract}

\section{Introduction}
\label{introduction}

\begin{figure}
	\centering
	\includegraphics[width =\columnwidth]{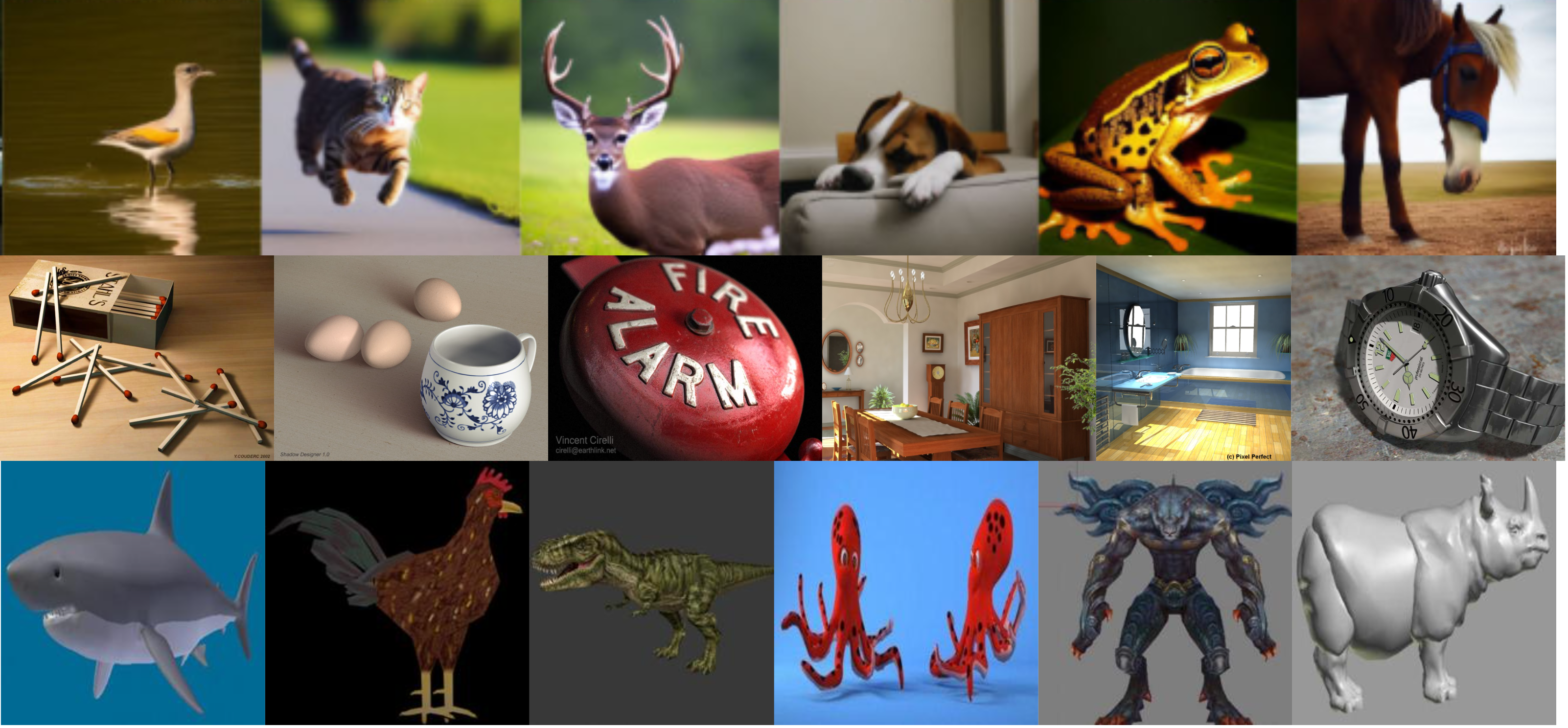}
	
	\caption{Examples of Computer Generated (CG) images from the CIFAKE-10, Columbia RCGI, and JSSSTU datasets (from top to bottom row, respectively). These images illustrate the challenge of distinguishing between CG images and natural images with the naked eye. The variation is also highlighted in computer-generated images across different datasets.}
	\label{fig1}
	
\end{figure}

The rapid advancement of computer-generated imagery (CGI) presents a significant challenge in distinguishing synthetic images from natural images (NIs) captured by digital cameras. The core challenges lie in the revolutionization of the computer graphics industry, provides the ability to produce synthetic images that convincingly replicate the authenticity of real-world scenes and animations. Some CGI can be so realistic that it is nearly indistinguishable from actual photographs, as shown in Figure \ref{fig1} with examples from the CIFAKE-10 \cite{bird2024cifake}, Columbia RCGI \cite{ng2005columbia}, and JSSSTU \cite{kumar2022dataset} datasets. As computer graphics continue to achieve increasingly photo-realistic results, the visual distinction between CGI and NIs becomes more subtle, posing significant challenges, particularly in fields like law and the judiciary.  For example, the CG can be used to temper with the actual evidence which will affect the normal judgment in the judicial system. Therefore, distinguishing between CG and NI has become a crucial topic in the field of digital forensics recent years.

Traditional methods for addressing this problem have typically been divided into subjective and objective approaches. Subjective methods often rely on human judgment through psychophysical experiments, whereas objective methods analyze images' statistical and intrinsic properties. Human observers excel in many multimedia applications, but the process can be both costly and time-consuming. Additionally, these evaluations are heavily influenced by various environmental factors and the individual’s ability to remain focused and attentive \cite{subjective}. Conventional objective techniques usually involve crafting sophisticated and discriminating features and applying classifiers such as support vector machines (SVM) or ensemble models \cite{lyu2005realistic,chen2007identifying,ng2005physics}. While these approaches may perform well on simpler datasets, they often fail when faced with more complex datasets that include images from a variety of sources.

Recent advancements in neural networks and vision transformers have revolutionized the field of computer vision, providing robust alternatives to traditional feature-based methods. Convolutional Neural Networks (CNNs) have shown remarkable capability in learning multi-level representations from data in an end-to-end manner, making them especially effective for complex classification tasks. Due to the huge success of CNNs, there is an increasing interest in applying these models to areas such as multimedia security and digital forensics. \cite{ni2019evaluation,carvalho2017exposing,quan2018distinguishing,yao2022cgnet}.

This research introduces a novel methodology for distinguishing between computer-generated and real digital images, leveraging the capabilities of Swin Transformers. Unlike traditional approaches, Swin Transformers eliminate the reliance on handcrafted feature extraction by directly processing raw pixel data, making them particularly effective for this classification task. Our method involves both intra-dataset and inter-dataset testing, utilizing RGB color space data to enhance classification accuracy. The approach is rigorously evaluated on three diverse datasets; CiFAKE, JSSSTU, and Columbia, demonstrating its robustness and generalization across different domains.
Our contributions are summarized as follows:
 
\begin{itemize} 
	\item We propose a Swin Transformer-based framework for differentiating CGI from real digital images, emphasizing intra-dataset and inter-dataset testing to evaluate robustness. 
	\item The model incorporates advanced preprocessing techniques in the RGB color space to enhance classification performance. 
	\item Our approach achieves state-of-the-art accuracy, consistently between 97-99\% across multiple datasets, demonstrating its efficacy in CGI detection. 
\end{itemize}

The remainder of this paper is structured as follows: Section \ref{introduction} outlines the challenge of distinguishing CGI from real images, emphasizing its importance in digital image forensics. Section \ref{relatedwork} reviews existing methods, from traditional techniques to recent profound learning advancements. Section \ref{method} details our proposed approach, including preprocessing steps and the Swin Transformer architecture used for classification. Section \ref{result} presents the experimental results, describing dataset usage, experimental setups, and performance evaluation. Finally, Section \ref{conclusion} concludes the paper with a summary of findings, a discussion of limitations, and suggestions for future research directions.

\section{Related Work}
\label{relatedwork}

Various methodologies have been developed in computer-generated (CG) image detection, primarily focusing on feature extraction and classification. One approach is to extract abnormal statistical traces left by specific graphic generation modules and employ threshold-based evaluation for detection. Ng et al. (2005b) \cite{ng2005physics} identified physical disparities between photographic and computer-generated images, achieving an 83.5\% classification accuracy by designing object geometry features. Wu et al. (2006) \cite{wu2006detecting} utilized visual features such as color, edge, saturation, and texture with the Gabor filter as discriminative features. Dehnie et al. (2006) \cite{dehnie2006digital} emphasized differences in image acquisition between digital cameras and generative algorithms, designing features based on residual images extracted by wavelet-based denoising filters. Texture-based methods have also been developed for CG and photographic (PG) classification. Li et al. (2014) \cite{li2014distinguishing} achieved 95.1\% accuracy using uniform gray-scale invariant local binary patterns. Peng et al. (2015) \cite{peng2015identification} combined statistical and textural features, enhancing performance with histogram and multi-fractal spectrum features. Despite their interpretability, hand-crafted feature-based methods are constrained by manual design limitations and feature description capacities.

In response to the limitations of hand-crafted features, recent research has leaned towards leveraging deep learning methods for improved detection performance. For instance,
Mo et al. (2018) \cite{mo2018fake} proposed a CNN-based method focusing on high-frequency components, achieving an average accuracy of over 98\%. Hu \textit{et al.}\cite{Hu2020} surveyed the research work done in distinguishing between realistic computer-generated (CG) and natural images (NI) in digital forensics, particularly using convolutional neural networks. Meena \textit{et al.} \cite{meena2021distinguishing} proposed a two-stream convolutional neural network, integrating a pre-trained VGG-19 network for trace learning and employing high-pass filters to emphasize noise-based features. Yao et al. (2022) \cite{yao2022cgnet} introduced a novel approach utilizing VGG-16 architecture combined with a Convolutional Block Attention Module, achieving an impressive accuracy of 96\% on the DSTok dataset. Furthermore, Chen et al. (2021) \cite{chen2021locally} presented an enhanced Xception model tailored for locally generated face detection in GANs. Liu et al. (2022) \cite{liu2022detecting} introduced a method focusing on authentic image noise patterns for detection. These recent advancements underscore the growing trend of applying sophisticated deep-learning architectures to enhance the detection accuracy of computer-generated images, addressing the challenges posed by increasingly realistic synthetic imagery. Gagan \textit{et al.} \cite{gagan2022} proposed a robust dual vision transformer approach operating in different color spaces, achieving 87\%-91\% accuracy in distinguishing natural images from both computer graphics and GAN-generated images. While these studies demonstrate various techniques for distinguishing between computer-generated and natural images, they are unable to provide a domain-generalized approach for identifying natural images across different types of synthetic imagery.

\section{Proposed Methodology}
This section presents the architecture of the Swin Transformer and includes feature visualization using $t$-SNE.
 
\label{method}
\label{methodology}
\begin{figure*}[!htbp]
	\centering
	
	\includegraphics[width=\columnwidth]{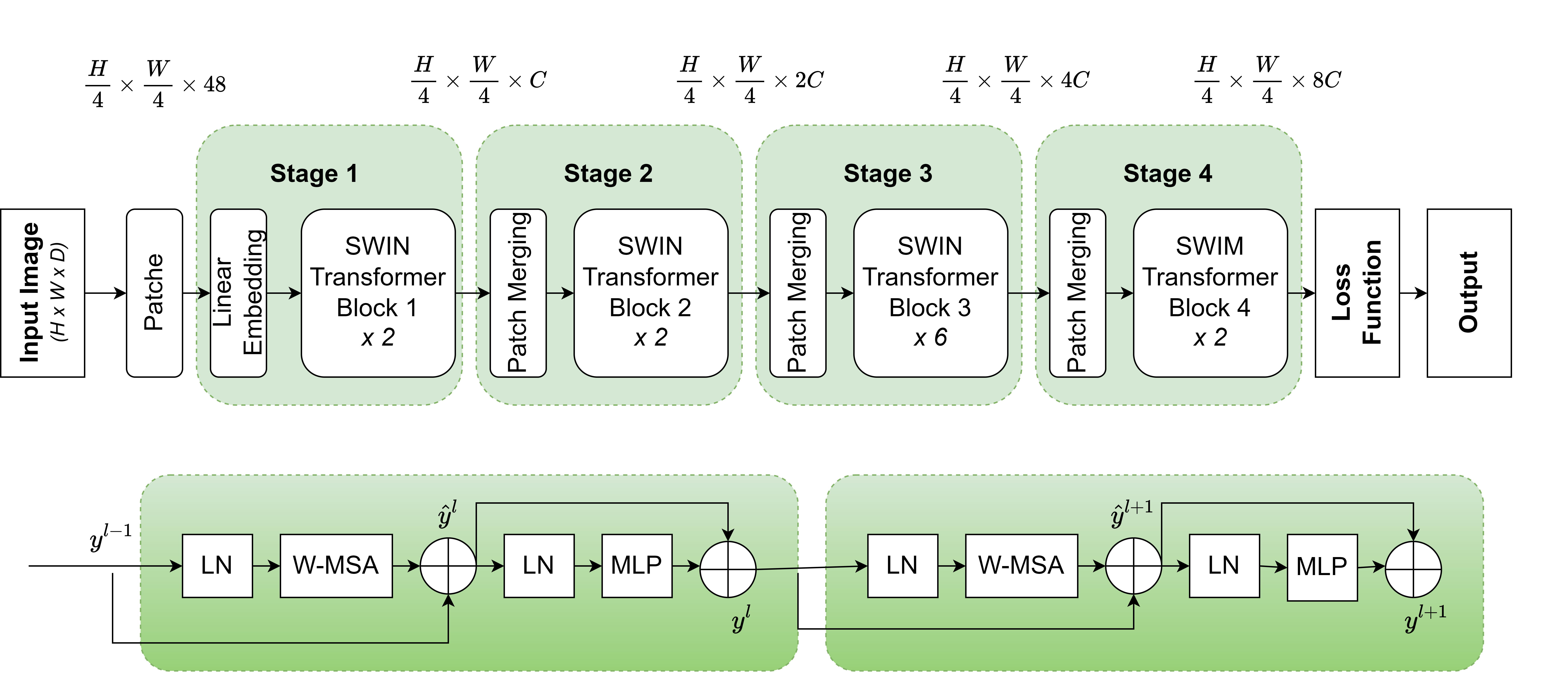}
	\caption{Architecture of the proposed Swin transformer and the expansion of Swin Transformer Block under it.}
	\label{fig2}
\end{figure*}

\subsection{Swin Transformer Architecture}
The Swin Transformer is a cutting-edge deep learning architecture renowned for effectively processing large-scale visual data with hierarchical representations. Unlike traditional convolutional neural networks (CNNs), the Swin Transformer adopts a hierarchical architecture that efficiently captures long-range spatial dependencies across the input images.

The architecture figure \ref{fig2} illustrates the hierarchical structure of the Swin Transformer, showcasing its ability to capture global context and local details simultaneously. By leveraging self-attention mechanisms and multi-layered processing, the Swin Transformer excels at learning complex patterns in visual data, making it ideal for image classification tasks.
It is characterized by self-attention mechanisms. Mathematically, the self-attention mechanism of the Swin Transformer can be represented as follow:

\[
\text{Att}(Q, K, V) = \text{softmax} \left( \frac{QK^T}{\sqrt{d_k}} \right) V
\]

where \( Q \), \( K \), and \( V \) represent the query, key, and value matrices, respectively, and \( d_k \) denotes the dimensionality of the key vectors.

Our study employed the Swin Transformer to distinguish between computer-generated images (CGI) and authentic images. We conducted color frame analysis to enhance the model's understanding of the distinct characteristics of CGI and authentic images. By analyzing the RGB color channels and extracting meaningful features, we aimed to provide the model with valuable insights into the color distribution and spatial arrangements between CGI and authentic images for all the three datasets.

\subsection{Feature Visualization with t-SNE}

To visualize the impact of color frame analysis on feature extraction, we employ t-Distributed Stochastic Neighbor Embedding (t-SNE) plots. Mathematically, t-SNE minimizes the Kullback-Leibler divergence between the distribution of high-dimensional feature vectors and their low-dimensional counterparts. The t-SNE algorithm can be represented as:

\[
p_{ij} = \frac{\exp(-\lVert x_i - x_j \rVert^2 / 2 \sigma_i^2)}{\sum_{k \neq l} \exp(-\lVert x_k - x_l \rVert^2 / 2 \sigma_k^2)}
\]
\[
q_{ij} = \frac{(1 + \lVert y_i - y_j \rVert^2)^{-1}}{\sum_{k \neq l} (1 + \lVert y_k - y_l \rVert^2)^{-1}}
\]
\[
C = \sum_i KL(P_i || Q_i) = \sum_{ij} p_{ij} \log \frac{p_{ij}}{q_{ij}}
\]

Here, \( p_{ij} \) represents the pairwise similarity between points \( x_i \) and \( x_j \) in the high-dimensional space, \( q_{ij} \) denotes the pairwise similarity between points \( y_i \) and \( y_j \) in the low-dimensional space, and \( C \) represents the Kullback-Leibler divergence.

\begin{figure}
	\centering
	\subfigure[\label{fig3a}]{\includegraphics[width=4cm,height=3cm]{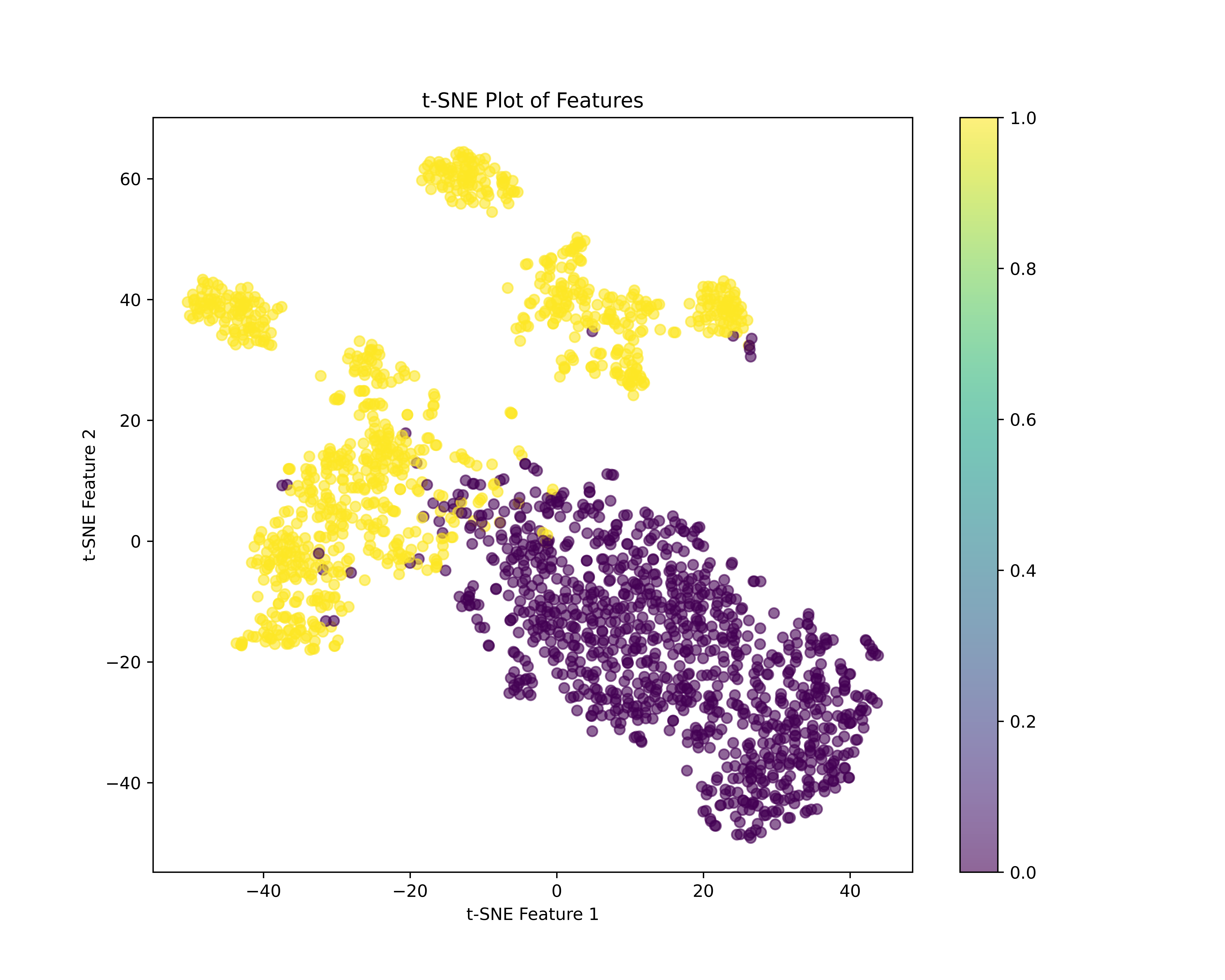}}
	\subfigure[\label{fig3b}]{\includegraphics[width=4cm,height=3cm]{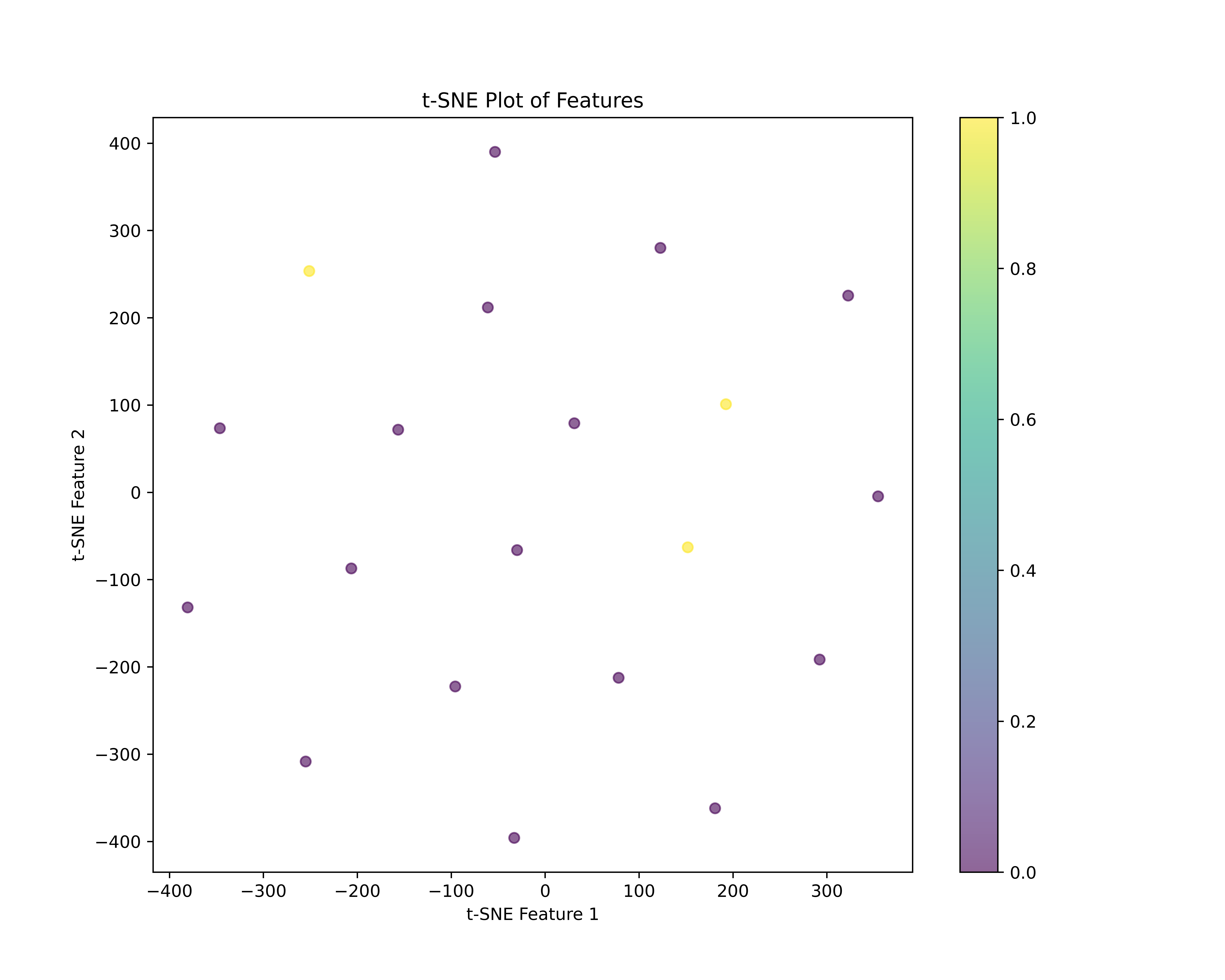}}\\
	\subfigure[\label{fig3c}]{\includegraphics[width=4cm,height=3cm]{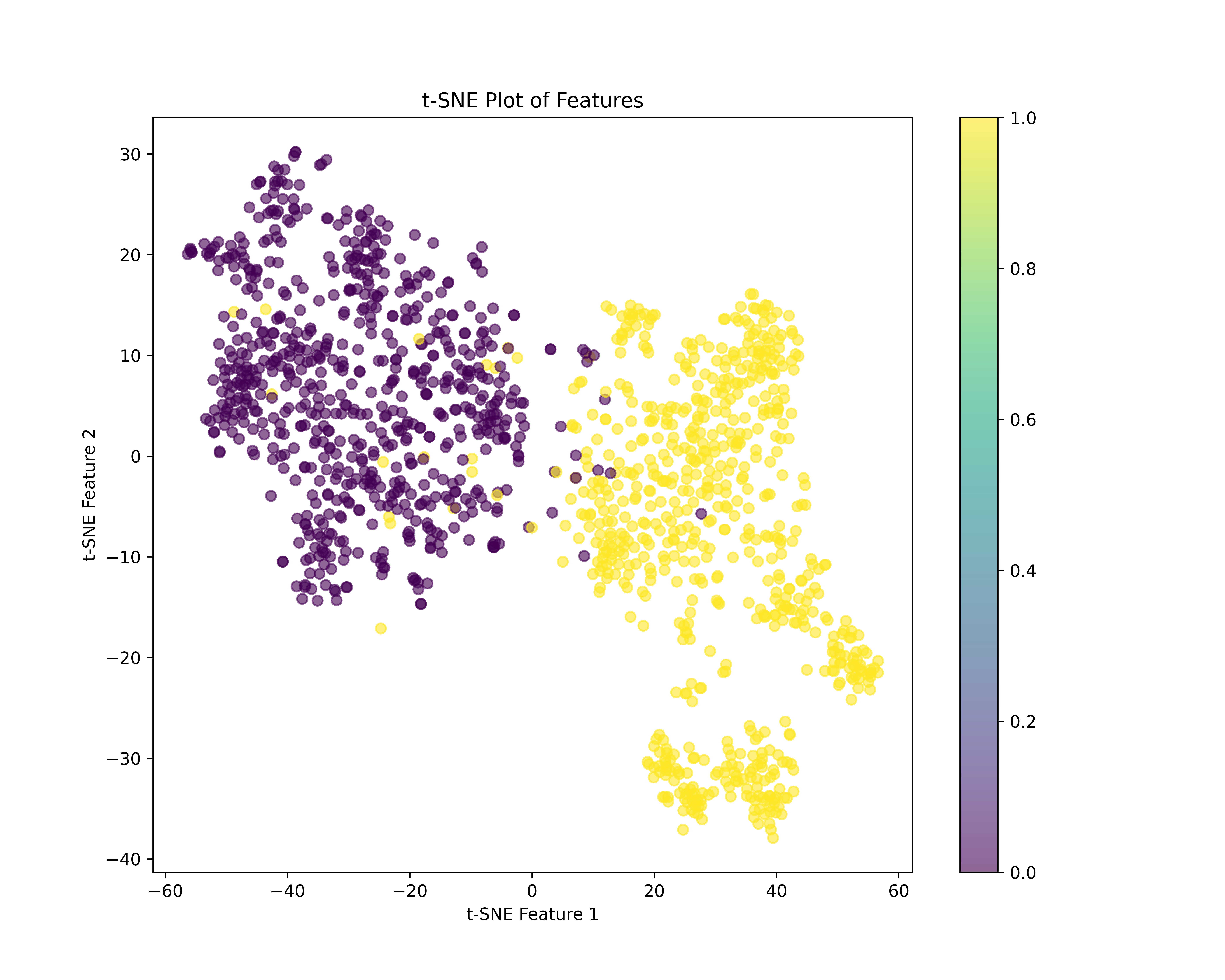}}
	\subfigure[\label{fig3d}]{\includegraphics[width=4cm,height=3cm]{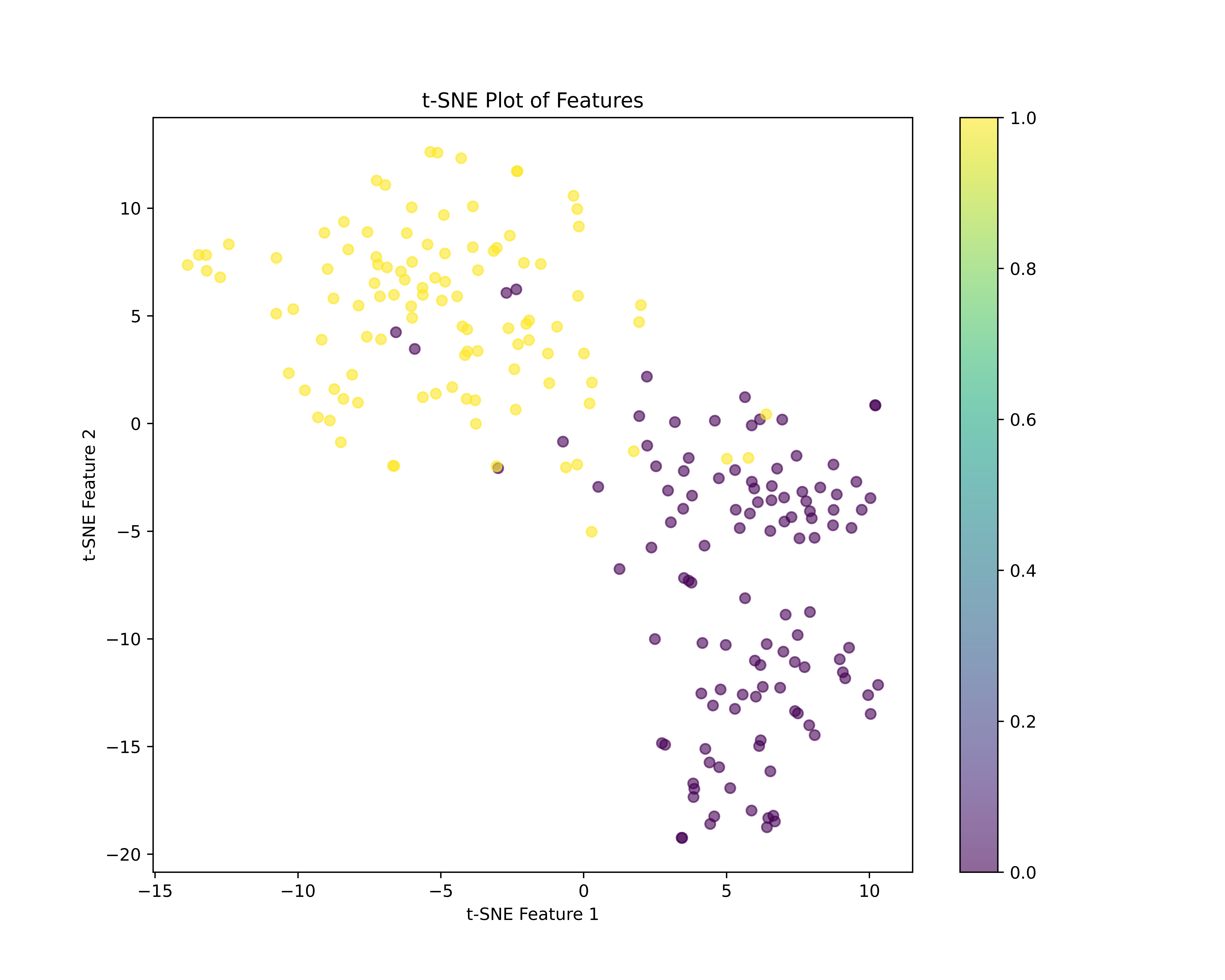}}
	\caption{The plots illustrate the t-SNE plot of the extracted features from the Swin Tranformer for the CiFake, columbia PRCG and real images, JSSSTU and combined all three dataset images (from left to right, top to bottom).  }
	\label{fig3}
\end{figure}

Figure \ref{fig3} showcases t-SNE plots of the extracted features, demonstrating the impact of color frame analysis on feature separability. The yellow dots represents CGI extracted features and purple dots represents the authentic image extracted features from the Swin Transformer in two-dimension space.

\section{Results and Discussion}
\label{result}

The Swin Transformer was trained utilizing features extracted from RGB color frames exclusively, owing to the characteristics of the datasets employed. The training objective centered on minimizing the cross-entropy loss function, defined as:
\[
\mathcal{L} = - \frac{1}{N} \sum_{i=1}^{N} \sum_{j=1}^{C} y_{ij} \log(p_{ij})
\]

where \( N \) is the total number of samples, \( C \) represents the number of classes, \( y_{ij} \) is the ground truth label, and \( p_{ij} \) is the predicted probability.

\subsection{Dataset Challenges and Curation}

The experimental setup involved three distinct datasets: CiFAKE (D1), Columbia (D2), and JSSSTU (D3). Each dataset comprised images classified into two categories: CGI and authentic images. However, we encountered significant challenges, particularly with the Columbia dataset. Numerous images from this dataset were corrupted or inaccessible due to broken URLs (HTTP 404 errors), substantially reducing the available image count. Specifically, only 43 CGI and 150 authentic images were retrievable from the intended 800 images per class.

\begin{table}[!htb]
	\centering
	\caption{Comparison of Dataset Attributes}
	\label{dataset_comparison}
	\resizebox{\columnwidth}{!}{
		\begin{tabular}{|l|p{4cm}|p{4cm}|p{4cm}|}
			\hline
			\textbf{Attribute} & \textbf{CiFAKE} & \textbf{Columbia} & \textbf{JSSSTU} \\ \hline
			\textbf{URL} & \href{https://www.kaggle.com/datasets/birdy654/cifake-real-and-ai-generated-synthetic-images}{Kaggle} & \href{https://www.ee.columbia.edu/~dvmmweb/dvmm/downloads/PIM\_PRCG\_dataset/dataset-download.htm}{Columbia University} & \href{https://sites.google.com/view/hrchennamma/research-activities/jssstu-data-sets}{JSSSTU} \\ \hline
			\textbf{Dataset Type} & Real \& AI-Generated Images & Photo-Realistic Computer Graphics (PRCG) \&Real & CGI \& Real Images \\ \hline
			\textbf{Image Count} & 1,000,000+ & 43 CGI, 150 Real (usable) & 14000 \\ \hline
			\textbf{Class Labels} & Fake, Real & PRCG, Google set & CG, PG \\ \hline
			\textbf{Image Resolution} & Varies & 1280x720 (typical) & Varies \\ \hline
			\textbf{Image Quality} & High-quality AI-generated and real images & Mixed quality, with some corrupted images & High-quality images \\ \hline
			\textbf{File Formats} & JPG & JPEG & JPG \\ \hline
			\textbf{Class Distribution} & Balanced & Imbalanced due to data loss & Balanced \\ \hline
	\end{tabular}}
\end{table}

To mitigate these discrepancies and ensure a balanced evaluation, we amalgamated the available images from all three datasets to form a unified dataset (D1+D2+D3). This composite dataset included 1500 images for each class (CGI and authentic), providing a robust basis for model training and evaluation. Table \ref{dataset_comparison} that contrasts the attributes of the three datasets—CiFAKE, Columbia, and JSSSTU.

\subsection{Model Evaluation Settings}

The performance of the proposed methodology was assessed using standard classification metrics: accuracy, precision, recall, and F1-score. These metrics are critical in quantifying the model's effectiveness in distinguishing between CGI and authentic images based on the extracted RGB features. The model was implemented in PyTorch and executed on a Dell Inspiron 5502, equipped with an Intel Core i5 processor and 16GB of RAM. Table \ref{table1} outlines the specific hyperparameters utilized during training.

\begin{table}[!htbp]
	\centering
	\caption{Hyperparameters}
	\label{table1}
	\begin{tabular}{|l|c|}
		\hline
		\textbf{Hyperparameters} & \textbf{Values}\\
		\hline
		Learning Rate & 0.0001\\ \hline
		Optimizer & Adam\\ \hline
		Loss Function & Cross Entropy\\ \hline
		Batch Size & 32\\ \hline
		Input Image Size & 224$\times$224$\times$3\\ \hline
		Normalized Mean & [0.485, 0.456, 0.406]\\ \hline
		Normalized Std & [0.229, 0.224, 0.225]\\ \hline
		Epochs & 5/10\\
		\hline
	\end{tabular}
\end{table}

The composite dataset (D1+D2+D3) derived from CiFAKE, Columbia, and JSSSTU datasets was carefully curated to maintain class balance. Despite the initial variability in image counts across datasets due to the abovementioned challenges, a balanced dataset was assembled with 1500 images per class (CGI and authentic). This approach allowed for a comprehensive evaluation of the model's performance across diverse data sources, ensuring the reliability and generalizability of the results.

\subsection{Simulation Results}

\begin{figure}[!htbp]
	\centering
	\subfigure[]{\includegraphics[width=3cm,height=3cm]{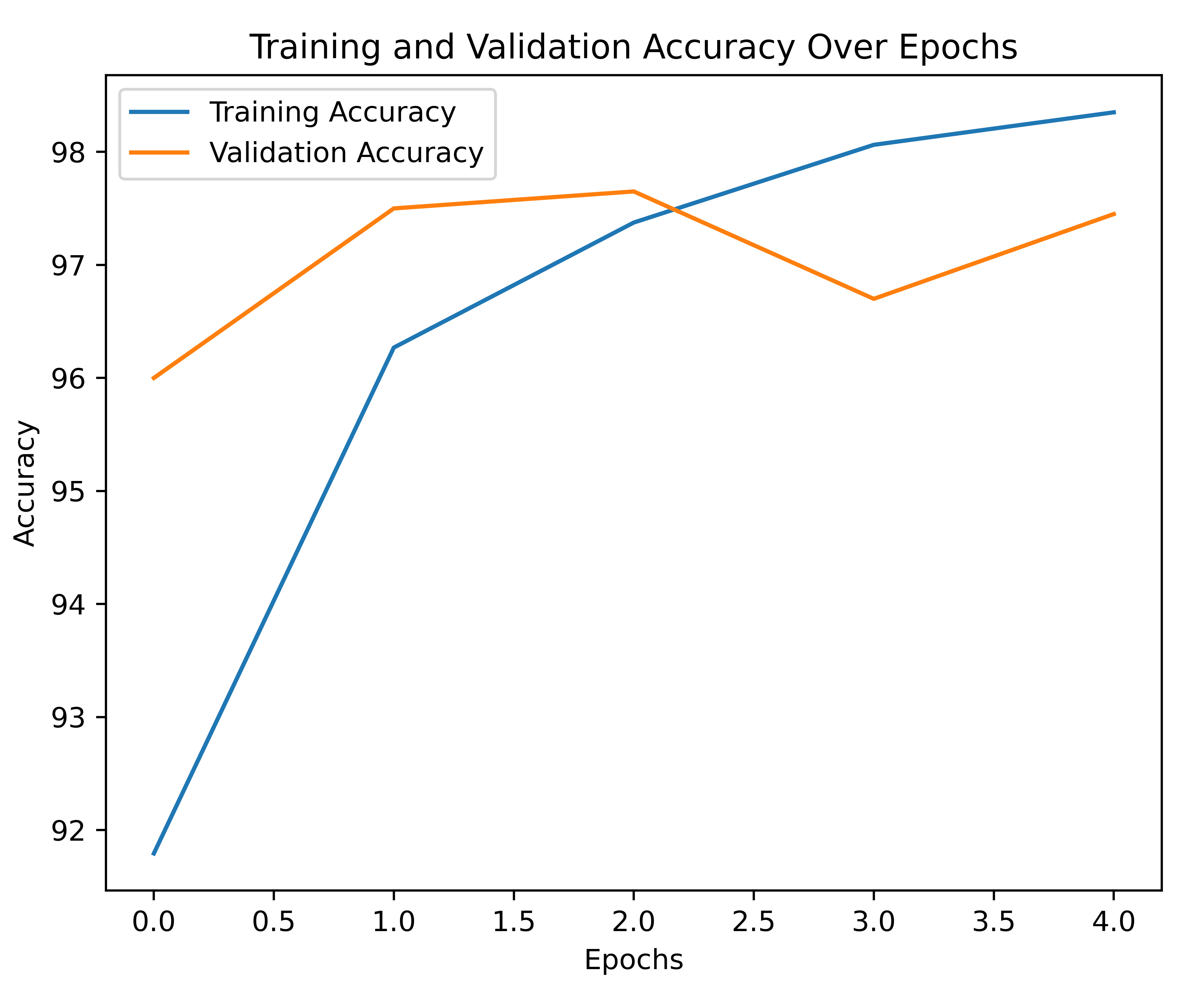}}
	\subfigure[]{\includegraphics[width=3cm,height=3cm]{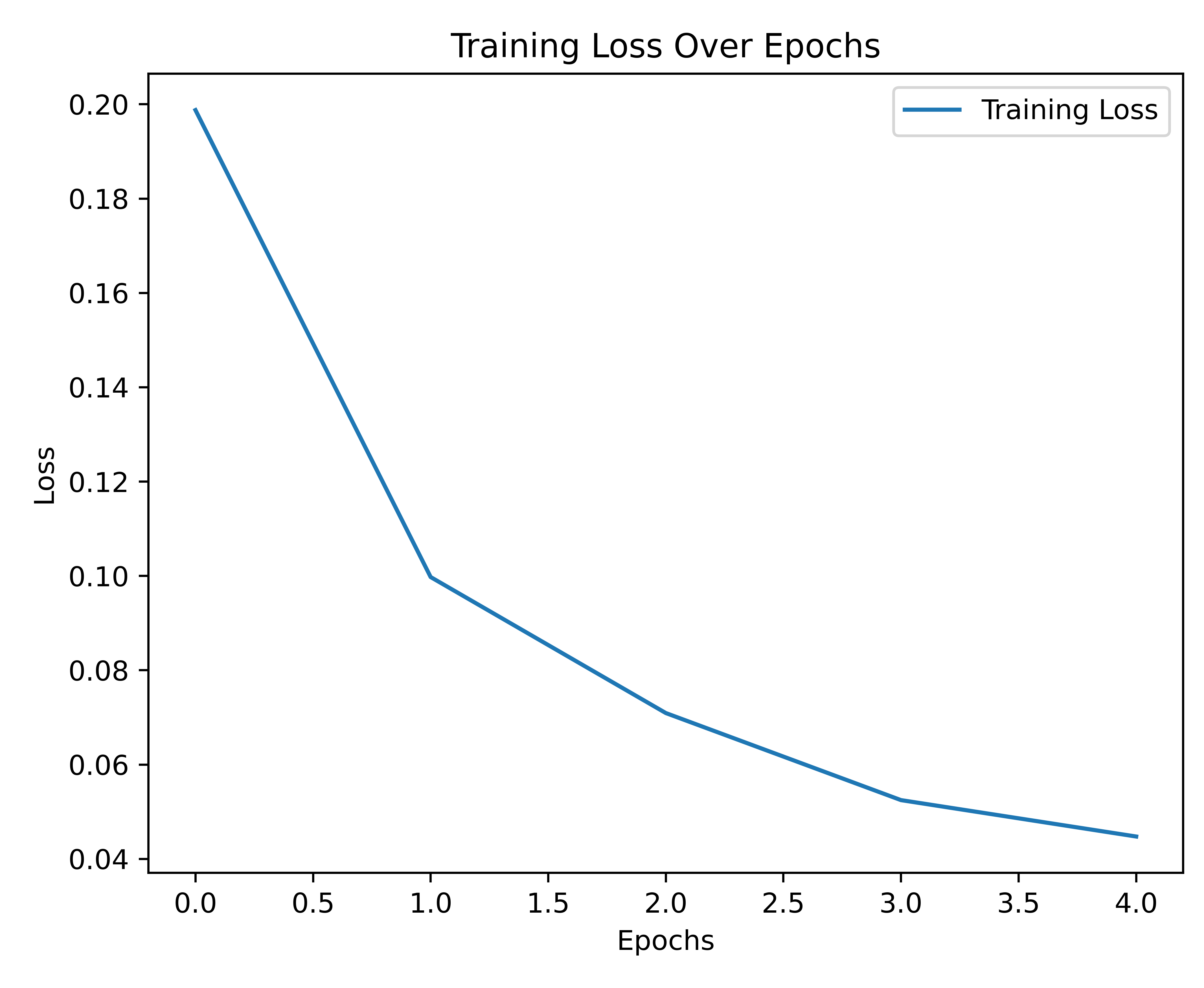}}
	\subfigure[]{\includegraphics[width=3cm,height=3cm]{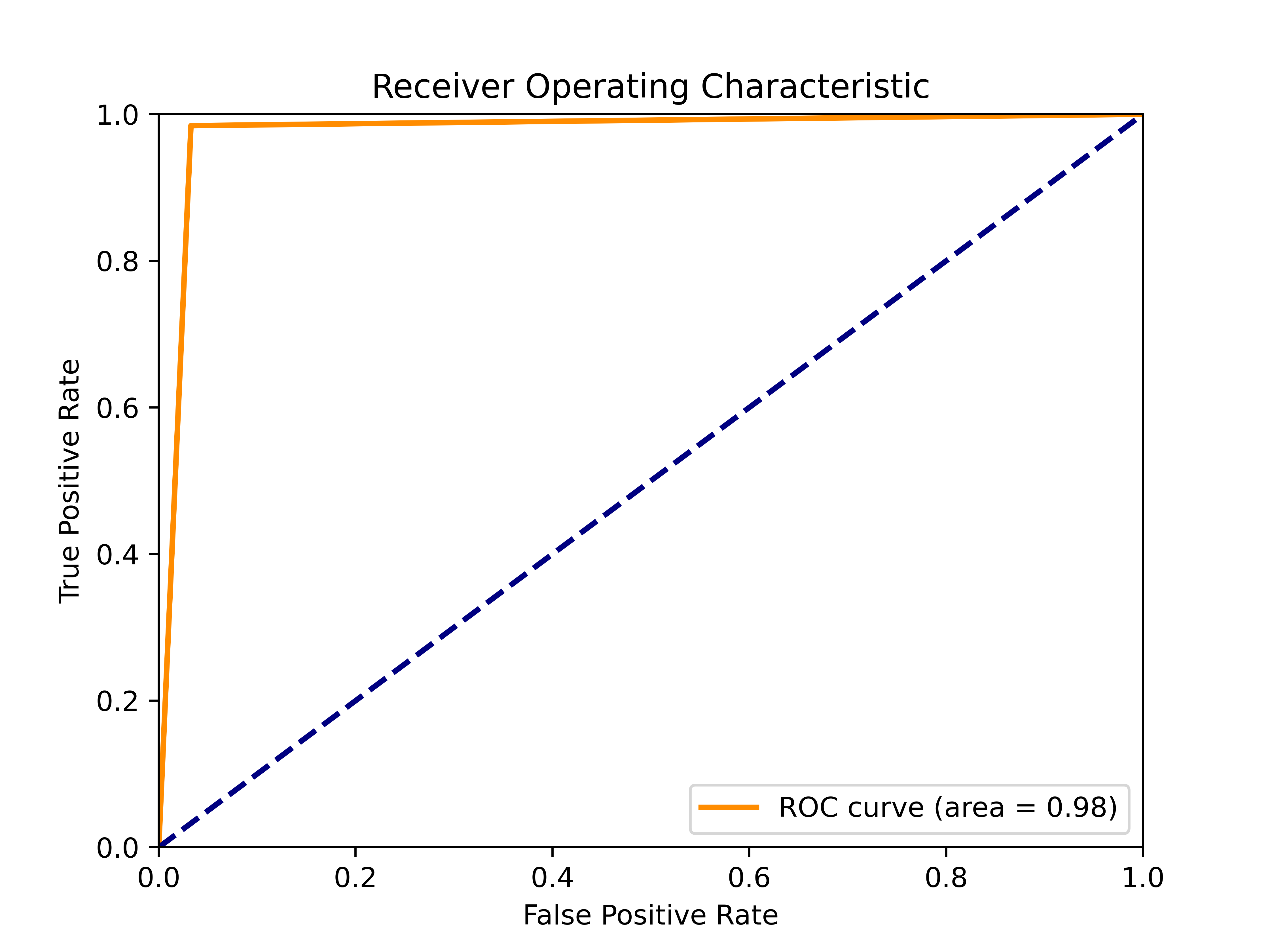}}\\
	
	\subfigure[]{\includegraphics[width=3cm,height=3cm]{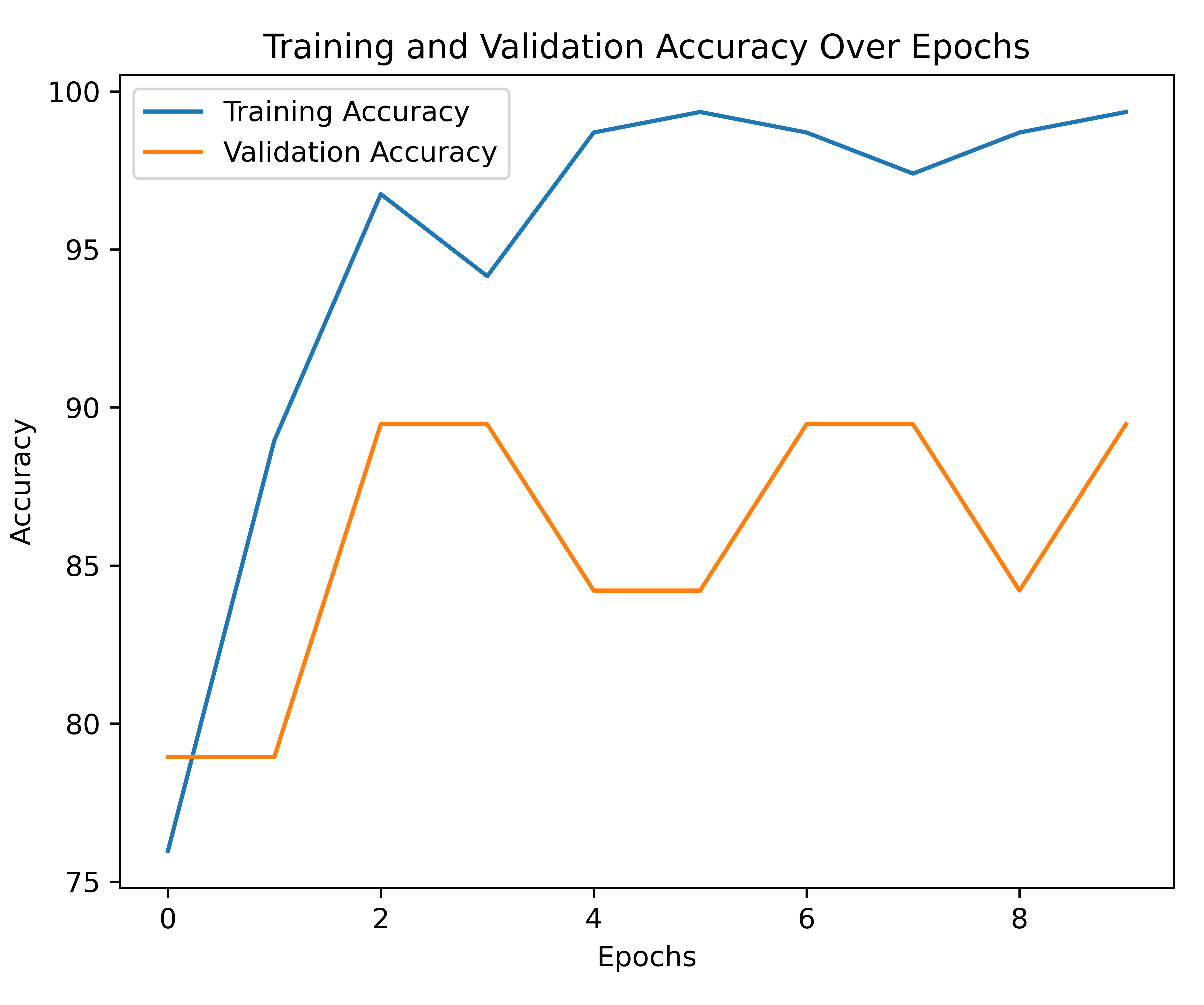}}
	\subfigure[]{\includegraphics[width=3cm,height=3cm]{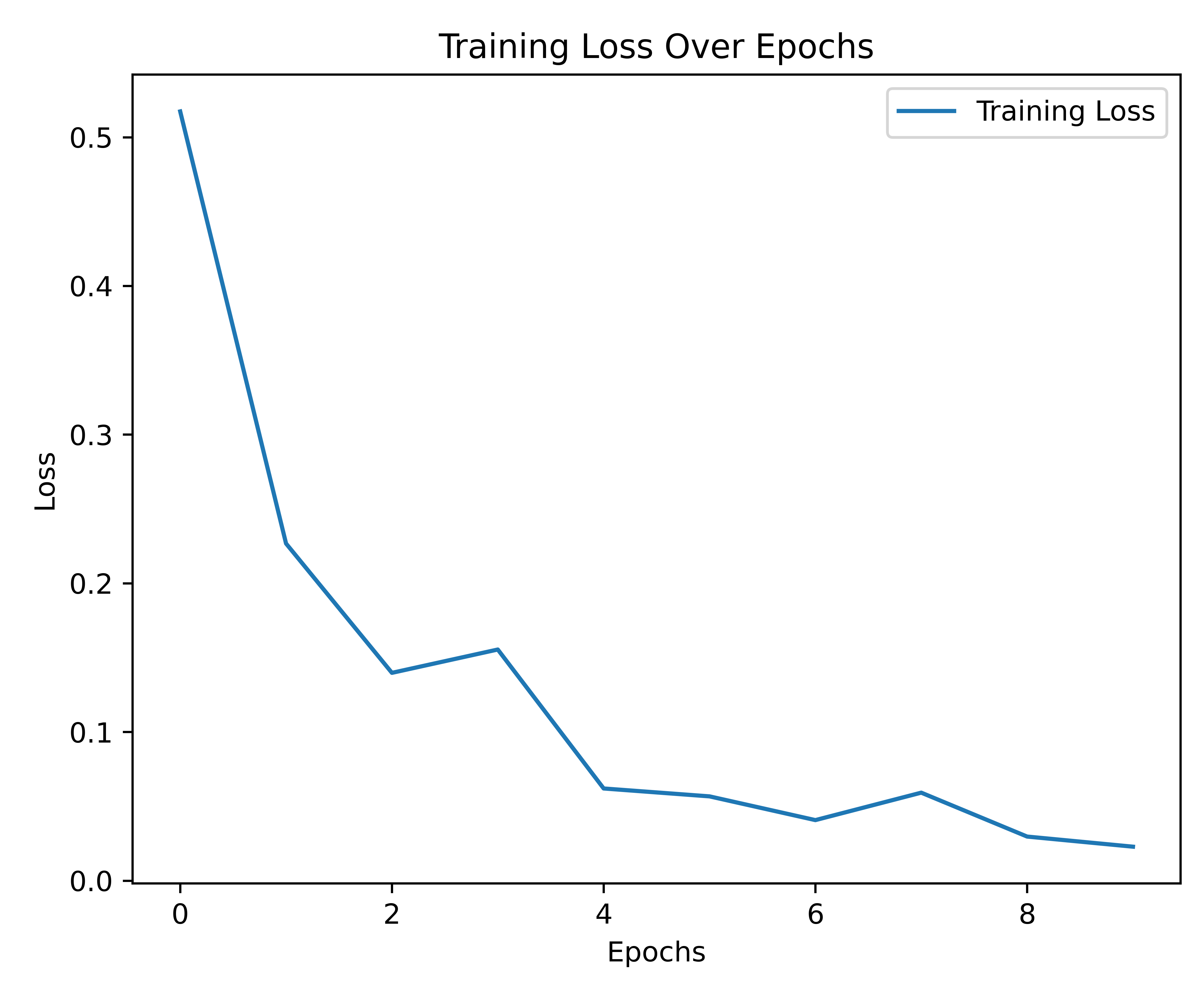}}
	\subfigure[]{\includegraphics[width=3cm,height=3cm]{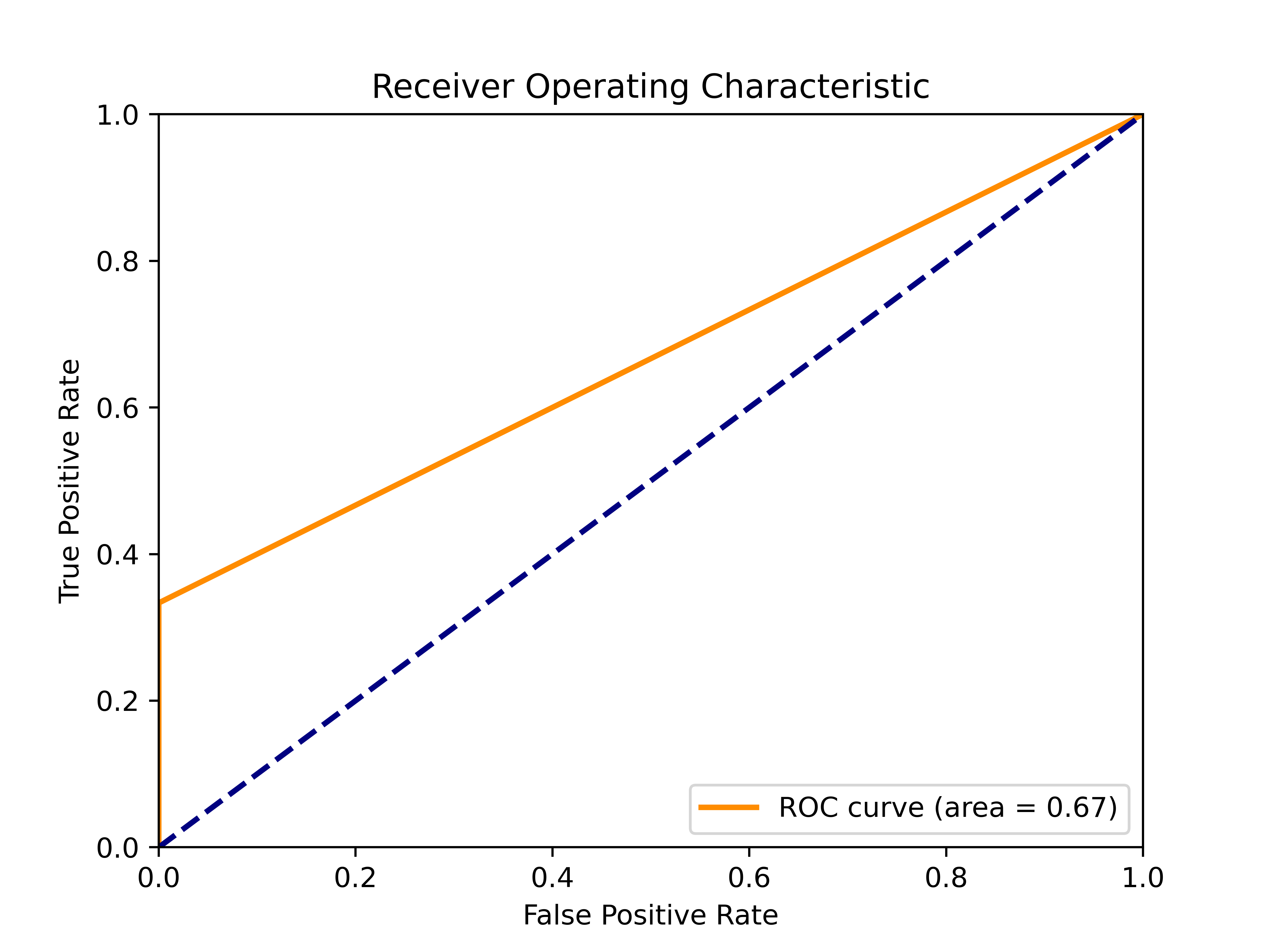}}\\
	
	\subfigure[]{\includegraphics[width=3cm,height=3cm]{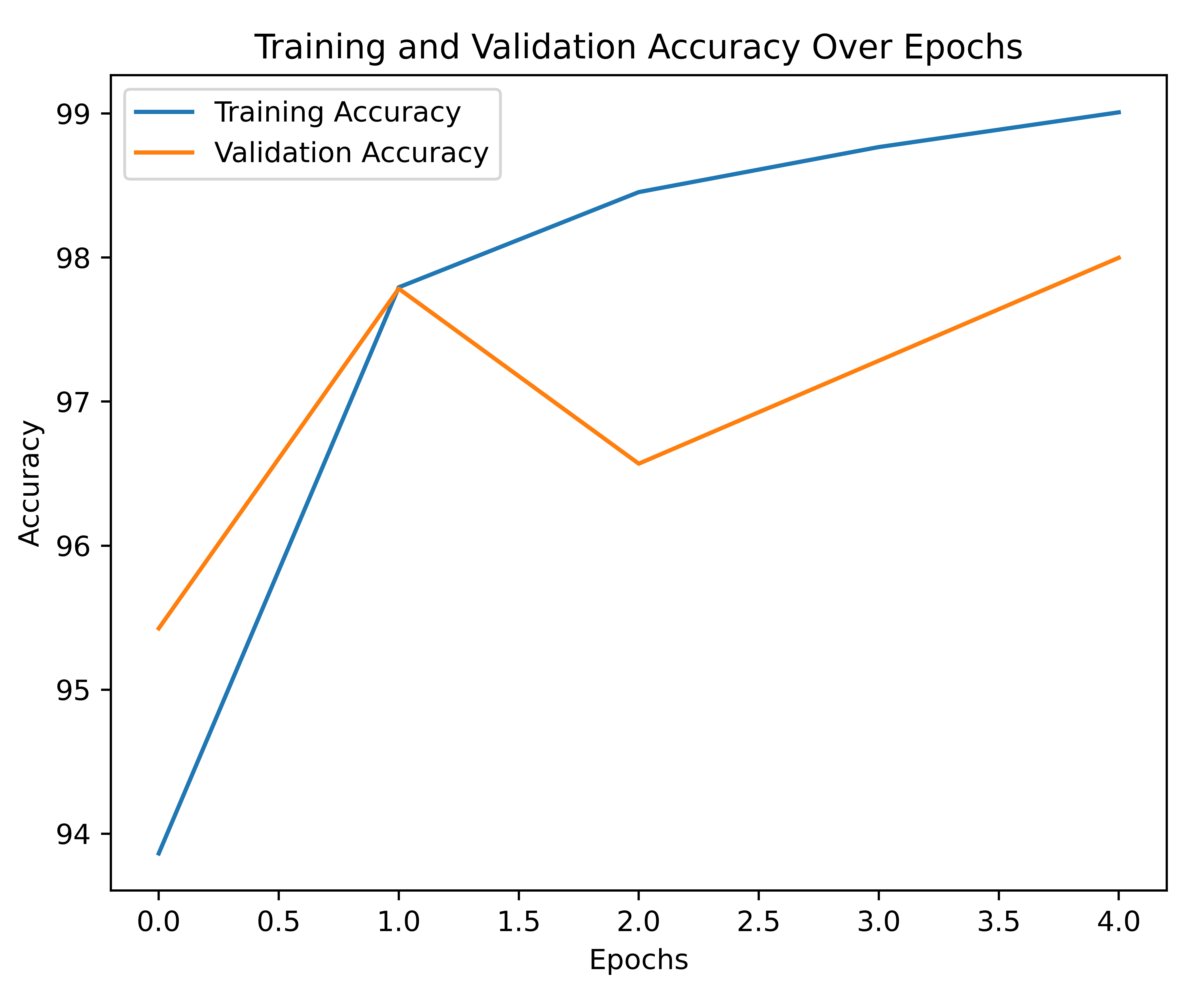}}
	\subfigure[]{\includegraphics[width=3cm,height=3cm]{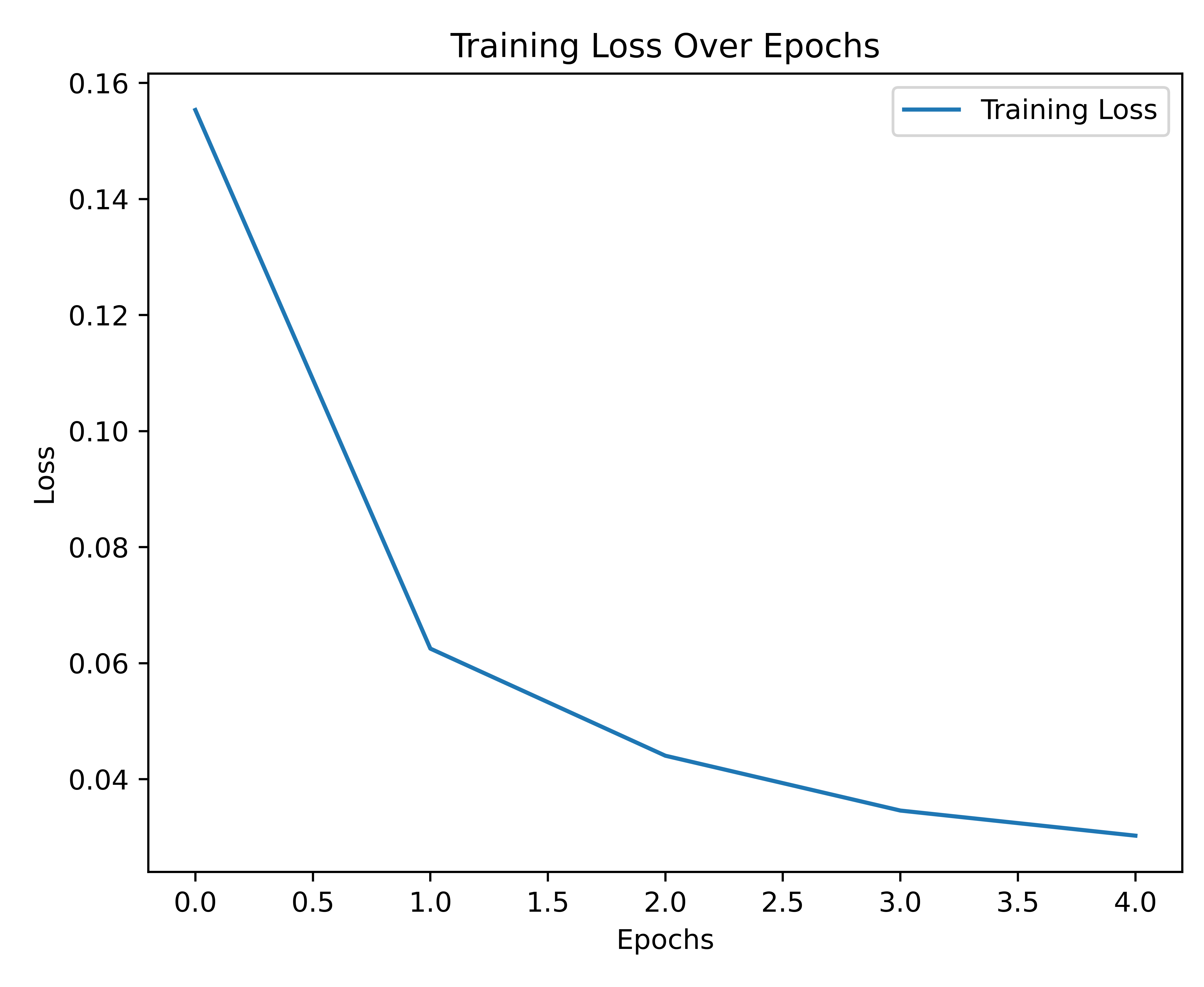}}
	\subfigure[]{\includegraphics[width=3cm,height=3cm]{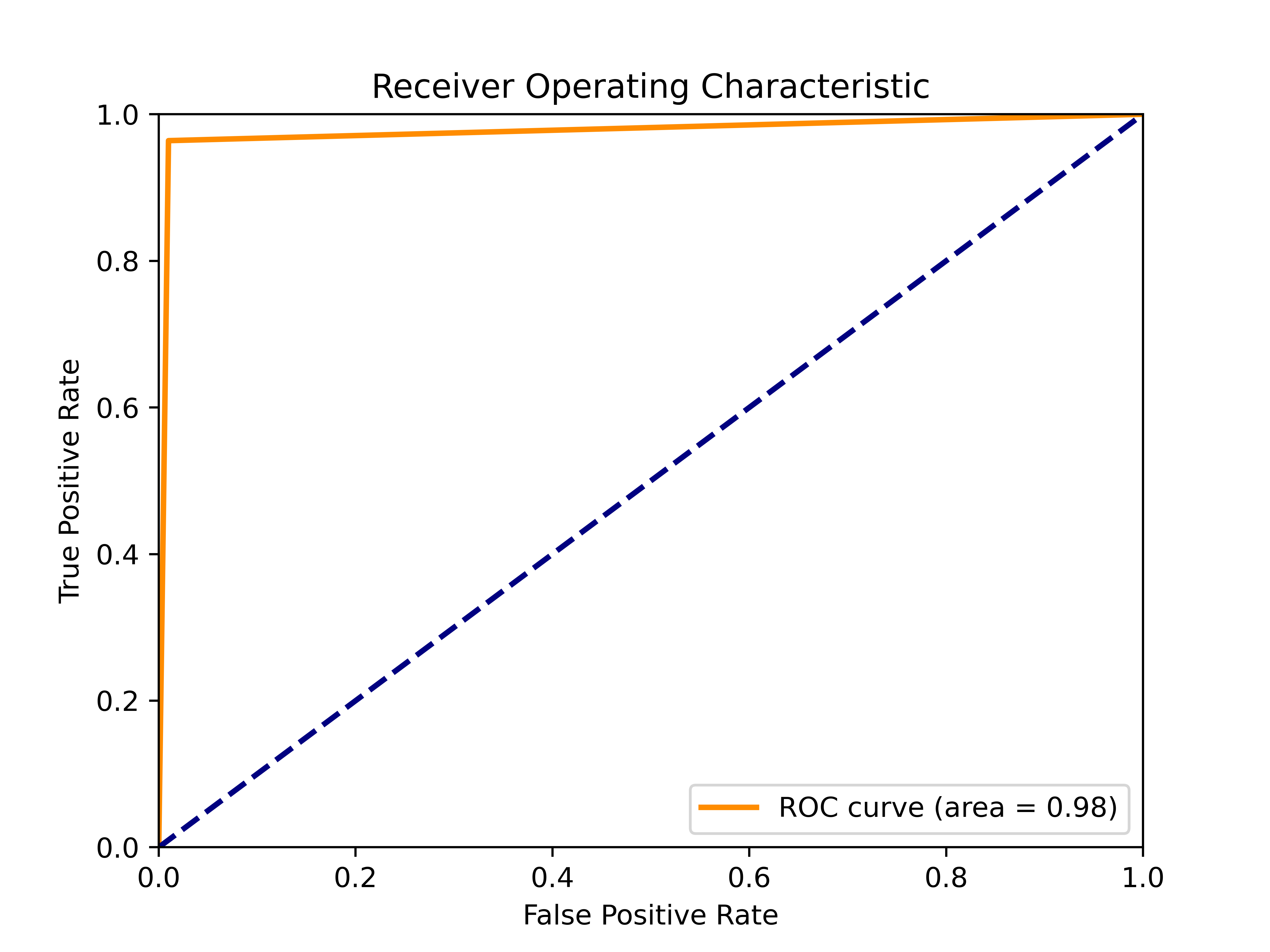}}\\
	
	\subfigure[]{\includegraphics[width=3cm,height=3cm]{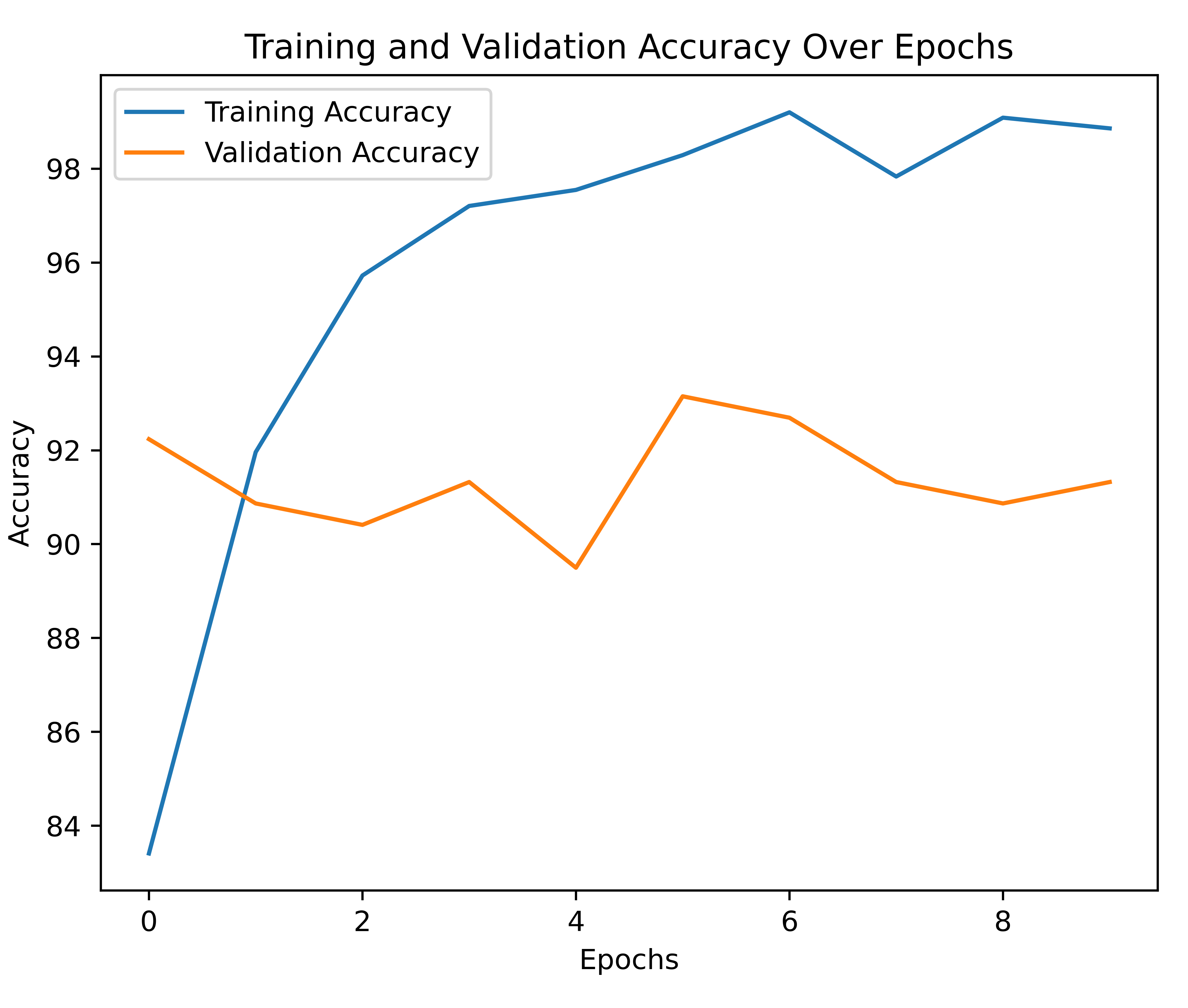}}
	\subfigure[]{\includegraphics[width=3cm,height=3cm]{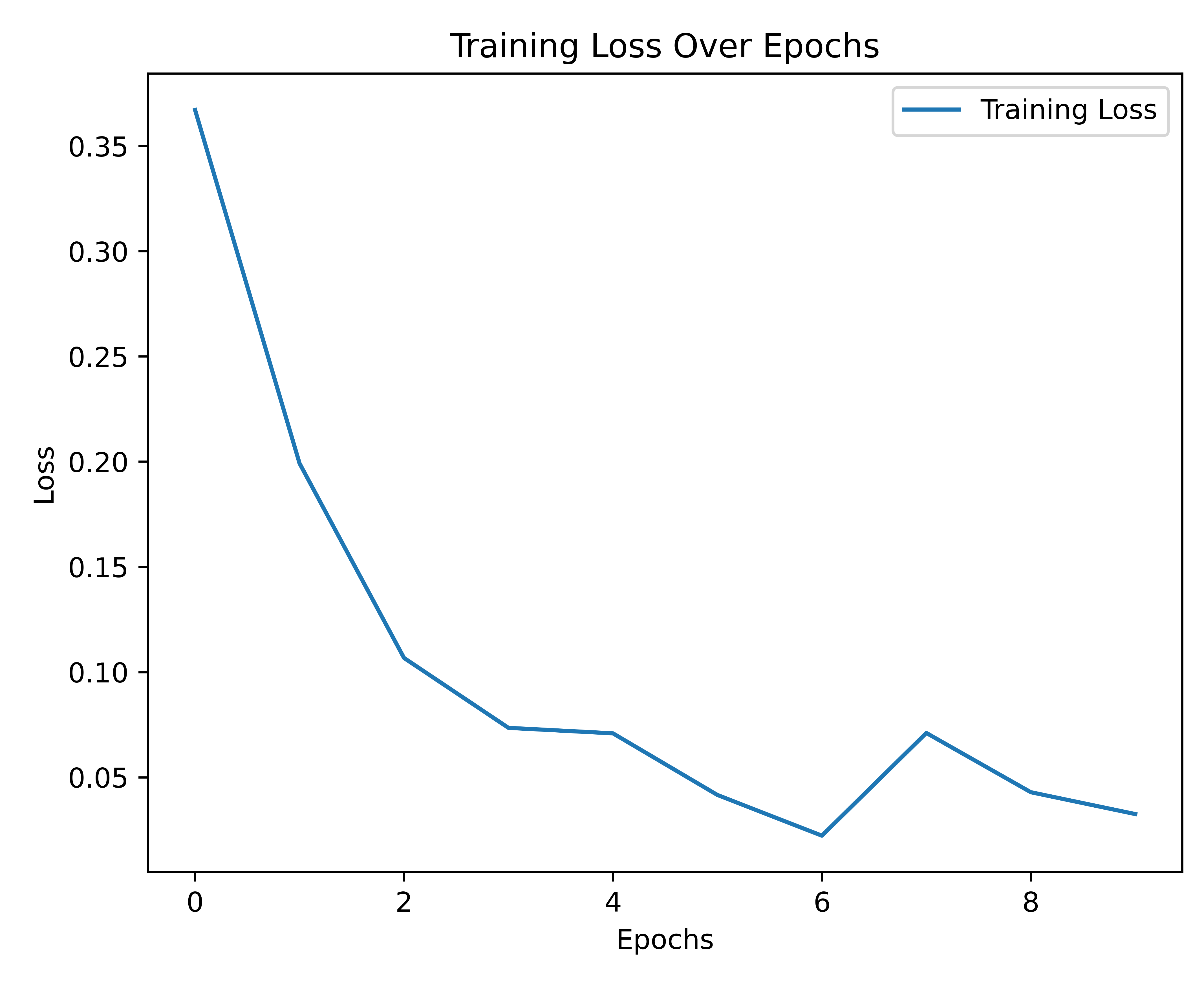}}
	\subfigure[]{\includegraphics[width=3cm,height=3cm]{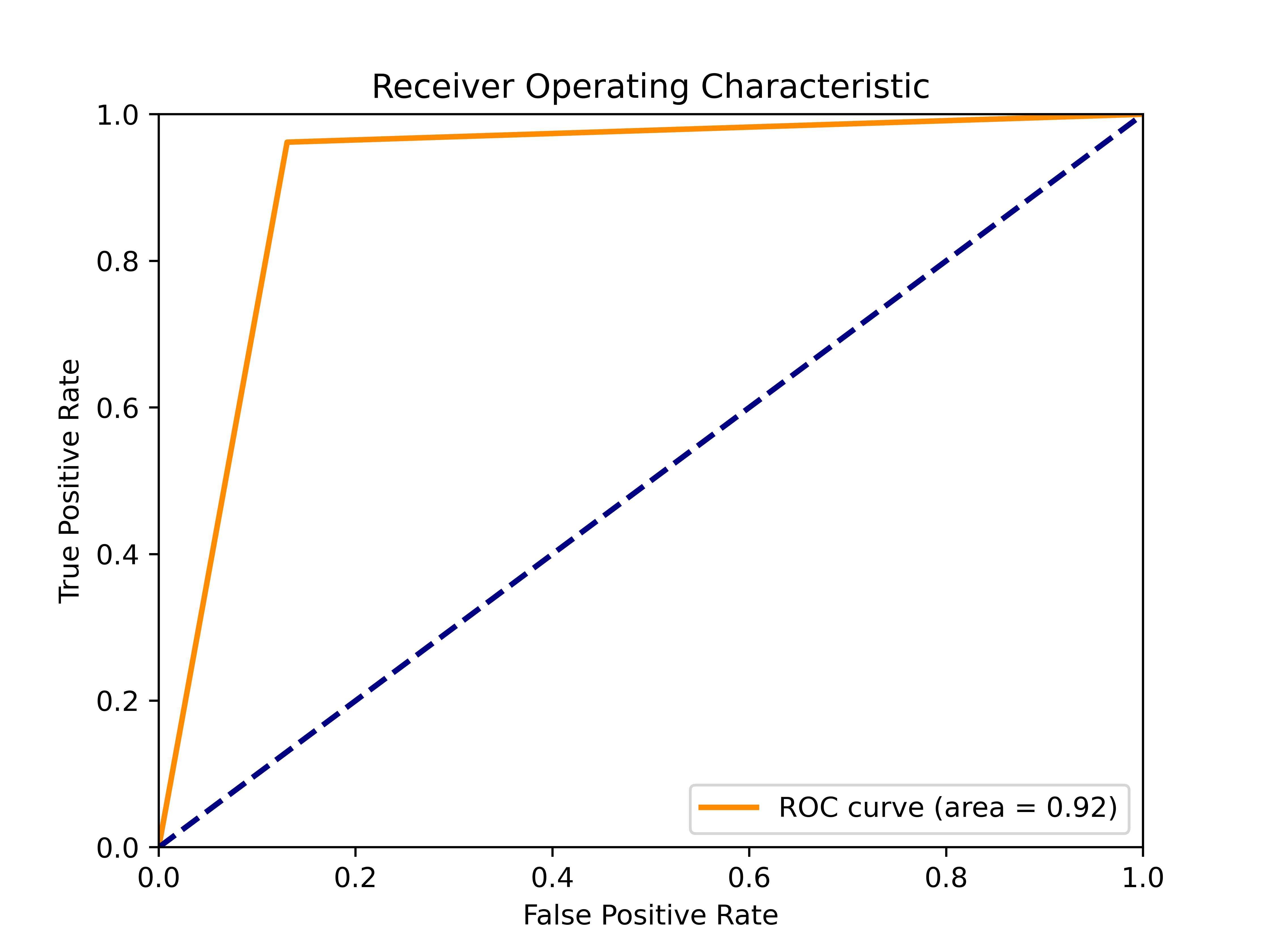}}
	
	\caption{The training and validation accuracy plot, loss plot and ROC (from left to right) for the datasets CiFAKE, columbia, JSSSTU, and combined three datasets (top to bottom).}
	
\end{figure}

The analysis of the results obtained from the Swin Transformer model, trained on the CiFAKE, Columbia, and JSSSTU datasets, reveals varying degrees of performance across these datasets. The training and validation accuracy curves, loss plots, and ROC curves provide insight into the model's effectiveness in distinguishing between CGI and authentic images. Each dataset presents unique challenges reflected in the model's performance metrics, such as accuracy, loss, and ROC AUC scores. The results demonstrate how the quality and quantity of data influence the model's learning process and its ability to generalize.

\subsubsection{Performance on CiFAKE and JSSSTU Datasets}

The CiFAKE and JSSSTU datasets exhibit relatively stable training processes, as indicated by the consistent increase in training accuracy and the smooth decrease in loss over the epochs. Despite showing a slight overfitting trend, the CiFAKE dataset maintains high training accuracy. However, the validation accuracy does not match up, suggesting that the model is overfitting to the training data. The JSSSTU dataset, on the other hand, shows a more balanced performance, with both training and validation accuracy improving steadily, indicating a better generalization capability. The ROC curve for the JSSSTU dataset, in particular, deviates more significantly from the diagonal, reflecting a more robust ability of the model to differentiate between CGI and authentic images in this dataset.

\subsubsection{Challenges with the Columbia Dataset}

The Columbia dataset presents significant challenges, as evidenced by the fluctuating validation accuracy and the relatively poor performance on the ROC curve. This can be attributed to the reduced dataset size after removing corrupted images, which likely resulted in a less representative sample of the overall data distribution. The instability in validation accuracy suggests that the model struggles to generalize from the limited and potentially biased training set. The dataset's smaller and more imbalanced nature leads to difficulties in learning robust features, making it harder for the model to distinguish between the two classes of images effectively.

\subsubsection{Combined Dataset Performance}

When combining the datasets into a single dataset (D1+D2+D3), the model's performance demonstrates the complexities of merging data from different sources with varying distributions and image qualities. The combined dataset shows instability during training, with fluctuations in validation accuracy and a slight increase in loss towards the end of the training process. The ROC curve for the combined dataset, while not as close to the diagonal as the Columbia dataset, still indicates that the model faces challenges in achieving high classification accuracy when dealing with data from diverse sources. 

\begin{table}[!htbp]
	\centering
	\caption{The evaluation parameters for the three datasets CiFake, columbia, JSSSTU and Combined.}
	\label{table 2}
	\begin{tabular}{|l|c|c|c|c|c|}
	\hline
	\textbf{Dataset} & \textbf{Accuracy} &\textbf{Precision} &\textbf{Recall} & \textbf{F1-score} &\textbf{AUC}\\ \hline
	\textbf{CiFake (D1)} &0.98 &0.97 &0.98 &0.97 &0.98\\
	\textbf{Columbia (D2)} &0.90 &1.00 &0.33 &0.50 &0.67\\
	\textbf{JSSSTU (D3)} &0.98 &0.99 &0.96 &0.98 &0.98\\
	\textbf{Combine (D1+D2+D3)} &0.91 &0.87 &0.96 &0.91 &0.92\\
	\hline
	\end{tabular}
\end{table}

In conclusion, the Swin Transformer model's performance varies significantly across the CiFAKE, Columbia, and JSSSTU datasets, with the Columbia dataset proving the most challenging due to its reduced and imbalanced nature. While the CiFAKE and JSSSTU datasets show more promising results, particularly in training stability and ROC curves, the combined dataset highlights the difficulties of generalizing across diverse data sources. These results underscore the importance of dataset quality and balance in training effective machine learning models and suggest that further refinements, such as advanced data augmentation, balancing techniques, or model tuning, may be necessary to improve performance across all datasets. The same can be seen in the evaluation parameters shown in table \ref{table 2}.

\section{Conclusion and Future Work}
\label{conclusion}

The experiments conducted on the CiFAKE, Columbia, and JSSSTU datasets using the Swin Transformer model revealed significant insights into the model's ability to distinguish between CGI and authentic images based on RGB color frame analysis. The model performed exceptionally well on the CiFAKE and JSSSTU datasets, achieving high accuracy and F1 scores, indicating its effectiveness in controlled settings. However, with limited and partially corrupted data, the Columbia dataset presented challenges that led to lower recall and AUC, highlighting the model's difficulties in generalizing under these conditions.
The combined dataset (D1+D2+D3) yielded moderate performance, giving insight into integrating high-quality and high-variance of integrating datasets with diverse characteristics for domain generalization

Future research could enhance the Swin Transformer model's robustness by incorporating additional data from more diverse sources to mitigate the challenges of imbalanced datasets. Exploring multi-modal approaches that combine RGB frame analysis with other feature types, such as texture or frequency domain features, could also provide more comprehensive input to the model, potentially leading to better classification accuracy. Furthermore, transfer learning, leveraging models pre-trained on more extensive and varied datasets, could also be explored to improve performance on smaller, more challenging datasets like Columbia. Finally, extending the analysis to include other color spaces, such as YCbCr, could offer additional insights into classifying CGI and authentic images.

\end{document}